\documentclass{article}

\usepackage[preprint]{neurips_2020}

\usepackage[utf8]{inputenc} 
\usepackage[T1]{fontenc}    
\usepackage{hyperref}       
\usepackage{url}            
\usepackage{booktabs}       
\usepackage{amsfonts}       
\usepackage{nicefrac}       
\usepackage{microtype}      
\usepackage{xcolor}         
\usepackage{graphicx}
\usepackage{subfig} 
\usepackage{comment}
\usepackage{amsmath}
\usepackage{multirow}
\usepackage{caption}
\usepackage{makecell}
\usepackage{bm}
\usepackage{comment}
\usepackage{array}
\setcitestyle{numbers}
\newcolumntype{x}[1]{>{\centering\arraybackslash\hspace{0pt}}p{#1}}
\setcitestyle{square}
\newcommand{\OURS}{Deep Distribution Transfer (DDT)}
\newcommand{\OURSABB}{DDT (ours)}
\newcommand{\OURSABBV}{DDT}
\title{Generalized Zero and Few-Shot Transfer for Facial Forgery Detection}

\author{%
  Shivangi Aneja\\
  Visual Computing Lab\\
  Technical University of Munich\\
  \texttt{shivangi.aneja@tum.de} \\
   \And
  Matthias Nie{\ss}ner \\
   Visual Computing Lab\\
   Technical University of Munich\\
   \texttt{niessner@tum.de} \\
}

\begin{document}

\maketitle

\begin{abstract}
We propose \OURS{}, a new transfer learning approach to address the problem of zero and few-shot transfer in the context of facial forgery detection. We examine how well a model (pre-)trained with one forgery creation method generalizes towards a previously unseen manipulation technique or different dataset. 
To facilitate this transfer, we introduce a new mixture model-based loss formulation that learns a multi-modal distribution, with modes corresponding to class categories of the underlying data of the source forgery method. 
Our core idea is to first pre-train an encoder neural network, which maps each mode of this distribution to the respective class labels, i.e., real or fake images in the source domain by minimizing wasserstein distance between them. 
In order to transfer this model to a new domain, we associate a few target samples with one of the previously trained modes. 
In addition, we propose a spatial mixup augmentation strategy that further helps generalization across domains.
We find this learning strategy to be surprisingly effective at domain transfer compared to a traditional classification or even state-of-the-art domain adaptation/few-shot learning methods. 
For instance, compared to the best baseline, our method improves the classification accuracy by 4.88\% for zero-shot and by 8.38\% for the few-shot case transferred from the FaceForensics++ to Dessa dataset.
\end{abstract}

\section{Introduction}

The rapid progress of image and video generation techniques has sparked a heated discussion regarding the authenticity and handling of visual content, for instance, on social media or online video platforms.
In particular, so-called {\em DeepFakes} videos have become a central topic, as they have shown the potential to create manipulated fake videos of human faces convincingly.
These methods include face-swapping techniques that artificially insert a specific person into a given target video~\cite{df_github,fsgan,petrov2020deepfacelab}, as well as facial reenactment methods where the goal is to modify the expression of a face such that the person appears to be saying something different~\cite{face2face,Kim_2018,thies2019neural}.
As a result, these methods could be misused, for instance, the speech of a politician or news commentator could be deliberately altered to communicate malicious propaganda backed with forged visual content to create a compelling forgery.

This increasing availability of manipulation methods calls for the need to reliably detect such forgeries in an automated fashion.
In particular, learning-based techniques have shown promising results on both images and videos proposed~\cite{ff_dataset,Nguyen_2019,face_warping,deepfake_inconsistent_head_pose,Zhou_2017,mesonet,agarwal_protecting_2019}.
While training supervised classifiers with one particular manipulation technique performs quite well on the same method, the main challenge lies in keeping the underlying training up to date with respect to the most recent forgery approaches.
For instance, the popular FaceForensics~\cite{ff_dataset} effort first provided a dataset of Face2Face~\cite{face2face} videos; a year later, Face Swap~\cite{fs_github} and DeepFakes~\cite{df_github} were added, and finally they authors released videos that were edited using NeuralTextures~\cite{thies2019neural}.
Unfortunately, providing constant updates for this supervised training is highly impractical since new manipulation techniques could appear from one day to another without any prior information.
This motivates us to re-think facial forgery detection from a zero and few-shot learning perspective.
Here, our aim is to generalize features between each method and reliably detect manipulations, despite having seen no or only very few training samples of a particular forgery technique.

Domain adaptation methods~\cite{tzeng2014deep,deep_coral,ccsa,dsne_2019} have the potential to be applicable in this scenario.
For instance, they aim to align classes from source and target domain irrespective of their original domain by learning a common subspace.
Other few-shot learning methods~\cite{relation_net_2017,snell2017prototypical,vinyals2016matching,finn2017modelagnostic,li2019revisiting} successively train a model on a large variety of classes with very few samples per class across different episodes, ultimately learning how to generalize for unseen classes. 
Both of these directions have been well-studied on the standard benchmarks with a larger number of classes and where samples from different classes are structurally very different from each other; e.g., Omniglot~\cite{lake2015human}, CUB~\cite{WelinderEtal2010}, miniImageNet~\cite{vinyals2016matching}, Office-31~\cite{officr_31}, VisDA~\cite{peng2017visda} etc..
However, we found that traditional domain adaption methods struggle in our forgery detection scenario, which maps to a binary classification problem with a high visual similarity between each class.

To address these challenges, we introduce \OURS{}, a novel transfer learning method tailored for forgery detection that encompasses both zero and few-shot learning to construct a domain-agnostic embedding space.
Specifically, we propose a new mixture model-based loss formulation that learns a multi-modal distribution, with modes corresponding to classes from the source domain. 
First, we pre-train an encoder network, such that the latent codes of each mode of the learned distribution are mapped to the respective class labels (Real and Fake) for the source dataset. 
To transfer this model to a new domain, we associate each sample from the target domain with one of the previously trained modes.
Classification is then performed by assigning the class label of the closest mode. 

To summarize, the key contributions in this paper are:
\begin{itemize}
    \item We propose \OURSABBV{}, a novel zero and few-shot transfer learning method for facial forgery detection method that explicitly models the underlying data with a multimodal distribution.
    \item We introduce a spatial mixup augmentation strategy that improves the model's generalization capability in unseen scenarios (w.r.t. manipulation methods and datasets).
    \item We provide an extensive analysis of transfer among different facial forgery methods within and across different datasets, showing that our approach outperforms state-of-the art on the Dessa dataset by 4.88\% in the zero-shot case and by 8.38\% for the few shot case.
\end{itemize}

\section{Related Work}

\textbf{Facial Manipulation Methods:}
Facial video manipulation methods have a long history in computer graphics, coupled with the 3D reconstruction of the underlying 3D face geometry~\cite{video_rewrite, Garrido_2014,face2face}.
Recently, we have also seen GAN-based synthesis methods generate high-quality facial imagery at remarkable detail~\cite{karras2019analyzing,karras2018stylebased, karras2017progressive, choi2017stargan, shen2019interpreting, y2019reconstructing, y2019reconstructing}; however, they struggle with temporal coherency.
Hybrid methods, such as Deep Video Portraits~\cite{Kim_2018} or Deferred Neural Rendering~\cite{thies2019neural}, combine the advantages of both directions by using rendered facial reconstructions as conditioning for generative neural networks, hence providing stable anchors for the temporal domain and enabling high-quality, photo-realistic video editing.
From an application standpoint, we can categorize these techniques into (a) facial re-enactment, which changes the expressions of the person while keeping the same identity, and (b) identity swapping, which replaces the facial region with another person's face.

\textbf{Facial Manipulation Detection:}
Traditional facial manipulation leverages handcrafted features such as gradients or compression patterns, in order to find inconsistencies within an image~\cite{jpeg_dimple, cfa_artifacts, region_splicing}.
While such self-consistency can produce good results, these methods are less accurate than more recent learning-based techniques based on convolutional neural networks~\cite{deepfake_inconsistent_head_pose}.
Hybrid methods can leverage both traditional and learned features in a two-stream fashion~\cite{Zhou_2017}, or focus on face-specific expressions and motion to identify an individual person~\cite{ff_dataset,agarwal_protecting_2019}.
These detection methods are, however, supervised in nature, and struggle to generalize between domains and datasets.

\textbf{Zero and Few-Shot Learning:} 
Zero and few-shot learning methods~\cite{chen2019selfsupervised, finn2017modelagnostic, vinyals2016matching, snell2017prototypical, relation_net_2017, garcia2017fewshot, rusu2018metalearning, qiao2017fewshot, Ravi2017OptimizationAA, li2019revisiting, kim2019variational, liu2018learning} have the potential to facilitate transfer across manipulation techniques or datasets; for instance, meta-learning based methods successively train a model on a large variety of tasks across different episodes.
An alternative direction is ProtoNets~\cite{snell2017prototypical}, which learns the prototype embedding for every class and uses the similarity between each prototype and query embedding for classification; or Relation Nets~\cite{relation_net_2017} which also computes the prototype for every class while proposing a learned similarity metric.
Very recently, Cozzolino et al.~\cite{forensictransfer} propose an autoencoder-based approach targeting forensics applications, including facial manipulations, which we consider an important baseline for our work.
Their method proposes to use a pre-trained embedding on the source domain, which is then fine-tuned on the target data.
While this leads to better results than a traditional binary cross-entropy loss, generalization nonetheless remains limited.
In contrast, we learn to model each class as a distribution, leading to significantly better generalization to new forgery methods as well as unseen datasets.

\section{Proposed Approach}

\begin{figure}[htpb]
  \centering
  \includegraphics[width=1.0\textwidth]{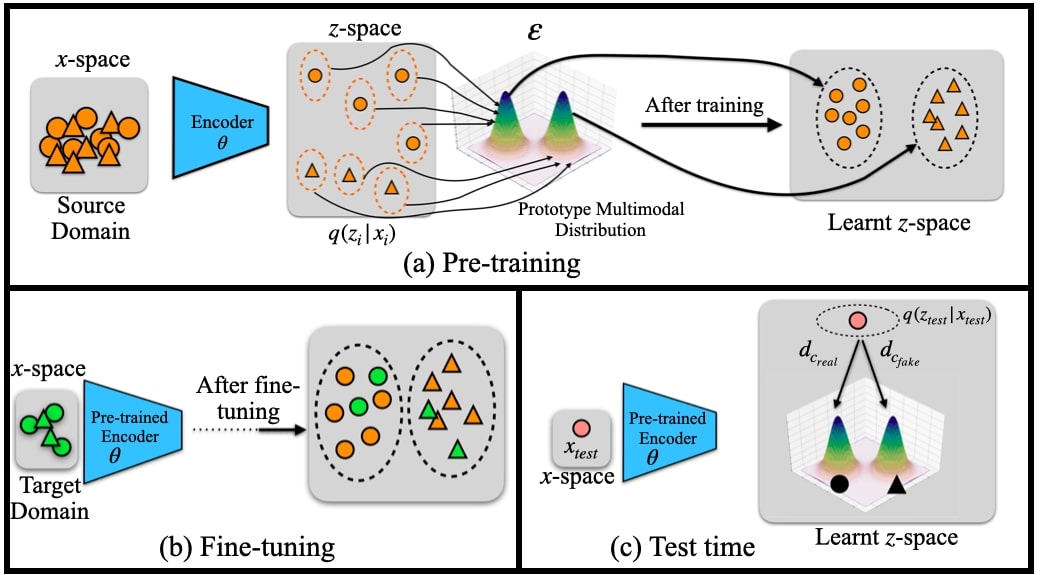}
  \caption{Method overview. (a) Pre-training: $\bm{\theta}$ encodes samples from the source domain into a latent distribution $q(\bm{z}_i|\bm{x}_i)$. $\bm{\epsilon}_c$ then maps class labels $c$ to the encoded distributions of the prototype multi-modal distribution $\bm{\varepsilon}$.
  (b) Fine-tuning: the pre-trained encoder $\bm{\theta}$ is used to map the few-shot samples from the target dataset with the same prototype multi-modal distribution $\bm{\varepsilon}$, which learns a common subspace between samples across domains. 
  (c) Test-time: a test sample $\bm{x}_{test}$ is encoded into the latent distribution $q(\bm{z}_{test}|\bm{x}_{test})$ by using the pre-trained encoder $\bm{\theta}$. We then compute the distance of the latent code with respect to all components of the distribution, and assign a class label based on the component it is closest to.}
  \label{fig:method_overview}
\end{figure}

The key idea of our method is to model classes as distributions, irrespective of their domain, where data points are projected into a learned embedding space where activated components reflect each activated class.
This multi-modal distribution can then be used to efficiently facilitate zero and few-shot learning in the respective target domain; see Fig.~\ref{fig:method_overview} for an overview.

\subsection{Modeling of Class Distributions}
We are given a large source dataset $\mathcal{S} = \{(\bm{x}_i,y_i)\}_{i=1}^N$ of $N$ labeled samples and a small target dataset $\mathcal{T} = \{(\bm{x}_i,y_i)\}_{i=1}^M$ of $M$ labeled samples such that $M \ll N$, where each $\bm{x}_i \in  \mathbb{R}^D$ is a $D$-dimensional input image and $y_i \in \{0,1 \}$ is the corresponding label ($0$ for Real, $1$ for Fake). 
$\mathcal{S}_c$ and $\mathcal{T}_c$ denotes the set of data points labeled with class $c$ in source and target dataset, respectively.

To model each class as a distribution, we learn a multi-modal Gaussian distribution $\bm{\varepsilon}$ with non-overlapping means, which enforces that every component $\bm{\epsilon}_c$ acts as a unimodal distribution in its own space.
By learning a single multi-modal distribution $\bm{\varepsilon}$ with non-overlapping means, we obtain many unimodal distributions $\bm{\epsilon}_c$, each representing a particular class $c$, which we refer to as a prototype multi-modal distribution.  
This distribution is a Gaussian mixture model consisting of several Gaussian distributions in the latent space, each identified by $ c \in \{0,...., C-1 \} $, where $C$ is the number of classes in our dataset.
Each Gaussian ${\bm{\epsilon}_c}$ in the mixture represents a class distribution and consists of following parameters:
\begin{itemize}
    \item Mean $\bm{m}_c$, defining its center.
    \item Covariance ${\Sigma}_c$, defining its width. For brevity, we assume $\Sigma_c = {I}$.
\end{itemize}
\begin{equation}
    {\bm{\varepsilon}}= 
\begin{cases}
    {\bm{\epsilon_0}} = \mathcal{N}{(\bm{z};\bm{m}_0,I)}\\
    {\bm{\epsilon_1}} = \mathcal{N}{(\bm{z};\bm{m}_1,I)}\\
    \vdots\\
    {\bm{\epsilon_{C-1}}} = \mathcal{N}{(\bm{z};\bm{m}_{C-1},I)}\\
\end{cases}
\label{eq:prot_mixture_model}
\end{equation}
Naively learning the class means $\bm{m}_c$ using the source training data $\mathcal{S}_c$ would severely overfit to specific forgery method or dataset; i.e., it would learn a specific feature representation for this dataset but without the capability to transfer to other manipulation methods or datasets. 
To this end, we regularize the model to ensure that the class means $\bm{m}_c$ do not overlap with each other.
Specifically, we constrain these means to be vectors consisting of 1s and 0s, with the equal number of neurons activating for each class. 
From the number of classes $C$ and size of the embedding space $K$, the class mean $\bm{m}_c$ with class label $c$ is then: 
\begin{equation}
    \bm{m}_c = \Big[ \big\{1\big\}_{i = p*(c...(c+1))}^{p}, \big\{0\big\}_{i \neq p*(c...(c+1))}^{p*(C-1)}  \Big]^K ,
\label{eq:class_mean}
\end{equation}
where $p$ is number of activated neurons per class and is given by $p= \frac{K}{C}$, with $i$ denoting the index location and $c$ the class label.

\subsection{Learning Class Distributions} 
Using an embedding function $\bm{\theta}$ with learnable parameters, we first project the data points from the $D$-dimensional input space ($\bm{x}$) to a smaller $K$-dimensional embedding ($\bm{z}$) space. 
Every data point $\bm{x}_i$ is modeled as a distribution $q(\bm{z}_i|\bm{x}_i) \sim\ \mathcal{N}(\bm{z}_i; \bm{\mu}_i, \Sigma_i)$  in this embedding space. 
To model each class as a single distribution component, we map all data point distributions $q(\bm{z}_i|\bm{x}_i)$ belonging to a particular class $c$ with a fixed component $\bm{\epsilon}_c$ of the prototype multi-modal distribution $\bm{\varepsilon}$ described above, given by Equation~\ref{eq:prot_mixture_model}.
For our binary classification case, this model learns to project each data point to a latent space $\bm{z}$, where each data point is mapped to only one component of a bimodal distribution ( $\mathcal{N}{(\bm{m}_0,I)}$ for \textit{Real} and $\mathcal{N}{(\bm{m}_1,I)}$ for \textit{Fake}); each component will then represent one of the classes.

\textbf{(1) Pre-training:}
Every sample $x_{i}$ in the source domain $\mathcal{S}$ is encoded into a latent distribution $q(\bm{z}_i|\bm{x}_i)$, where we minimize the distribution divergence metric $d$ between this encoded latent distribution $q(\bm{z}_i|\bm{x}_i)$ with the corresponding component of our fixed prototype multi-modal distribution $\bm{\epsilon}_c$ (based on its class label $c$): 
\begin{equation}
    \mathcal{L}_{pretrain} = \frac{1}{N} \Bigg[ \sum_{i=1}^N d \Bigg(q(\bm{z}_{i}|\bm{x}_{i}, {y}_{i}), \bm{\epsilon}_c\Bigg) \Bigg]
\end{equation}
such that $y_i = c$, where $c$ is the ground truth label for sample $\bm{x}_i$ and $\bm{\epsilon}_c$ is the corresponding prototype distribution component.
This minimization learns an embedding subspace such that all samples belonging to a class $c$ are aligned close to each other with the component $\bm{\epsilon}_c$ of the prototype distribution; samples belonging to different classes are mapped with the separate components (see Fig.~\ref{fig:method_overview}(a)).

\textbf{(2) Fine-tuning (few-shot only):}
For the target domain $\mathcal{T}$, we associate the few available training samples with the same prototype distribution $\bm{\varepsilon}$ used before for learning the source domain $\mathcal{S}$, according to its class label.
We then fine-tune the above pre-trained model $\bm{\theta}$ with the target dataset.
In order to further mitigate overfitting during fine-tuning, we propose a spatial mixup augmentation strategy.
Specifically, we spatially mix two images in a vertical fashion, forming a new image by taking half of the face from one image and another half from the other images.

\textbf{(3) Testing:}
We encode each test sample $\bm{x}_{test}$ into the latent distribution $q(\bm{z}_{test}|\bm{x}_{test})$ by using the trained encoder $\bm{\theta}$. 
We then compute the distance of this encoded latent distribution with respect to all the components of the prototype multi-modal distribution, and assign the label to the component it is closest to: 
\begin{equation}
    \hat{y}_{test} = argmin\Big\{d_0, d_1,.... d_{C-1}\Big\}
\end{equation}
where $d_c$ is the distribution divergence distance between $q(\bm{z}_{test}|\bm{x}_{test})$ and $\bm{\epsilon}_c$.

\textbf{Distribution Alignment Distance:}
Every sample $\bm{x}_i$ that is fed to encoder model $\bm{\theta}$ outputs a distribution $q(\bm{z}_i|\bm{x}_i) \sim \mathcal{N}(\bm{z}_i;\bm{\mu}_i, \bm{S}_i)$. To align this distribution with the class component $\bm{\epsilon}_c$ of the prototype distribution, we minimize the Wasserstein distance between them.

In case of multivariate Gaussians, a closed-form solution of the 2-Wasserstein distance~\cite{w_gauss} between two distributions $\mathcal{P}$ and $\mathcal{Q}$ is given by:
\begin{equation}
     W_{\mathcal{P}\mathcal{Q}} = \Bigg[ \| \mu_\mathcal{P} -\mu_\mathcal{Q} \|_2^2 + Tr\big[ \Sigma_\mathcal{P} + \Sigma_\mathcal{Q} - 2\big(\Sigma_\mathcal{P}^{\frac{1}{2}}\Sigma_\mathcal{Q}\Sigma_\mathcal{P}^{\frac{1}{2}}\big)^{\frac{1}{2}} \big] \Bigg]^{\frac{1}{2}} 
    \label{eq:wd_1}
\end{equation}
In our case, we assume that encoder $\bm{\theta}$ predicts diagonal covariance matrix, which allows us to simplify the distance:
\begin{equation}
     d(q_i, \bm{\epsilon}_c) = W_{q_i\bm{\epsilon}_c} = \Bigg[ \| \bm{\mu}_i -\bm{m}_c \|_2^2 +  \|\bm{S}_i^{\frac{1}{2}} - \bm{I}^{\frac{1}{2}}\|_{\text{Frob}}^{2}  \Bigg]^{\frac{1}{2}} 
    \label{eq:wd_2}
\end{equation}

\section{Datasets}

\begin{figure}[htpb]
    \centering
    \subfloat[\small{FF++~\cite{ff_dataset}}]{{\includegraphics[width=0.19\textwidth]{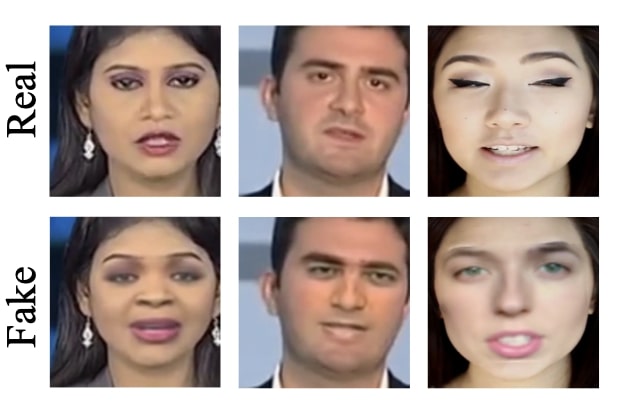} }}%
    \hfill
    \subfloat[Google DFD~\cite{dfdc_google}]{{\includegraphics[width=0.19\textwidth]{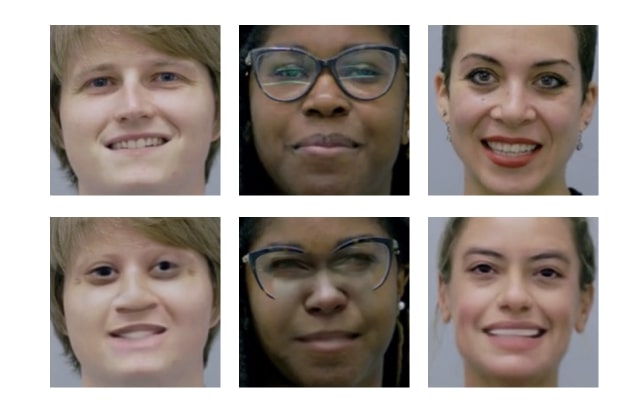} }}%
    \hfill
    \subfloat[Dessa~\cite{dessa_dfdc}]{{\includegraphics[width=0.19\textwidth]{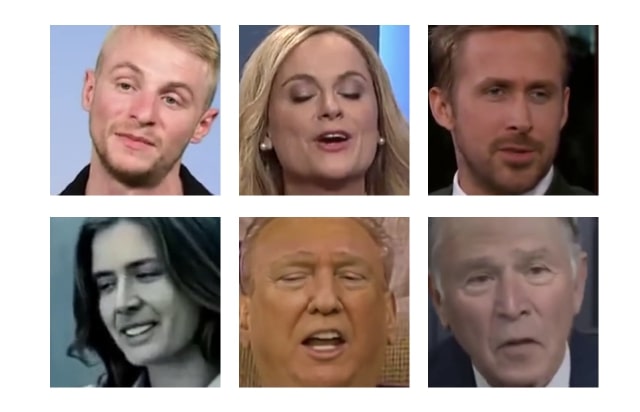} }}%
    \hfill
    \subfloat[Celeb DF~\cite{Celeb_DF_cvpr20}]{{\includegraphics[width=0.19\textwidth]{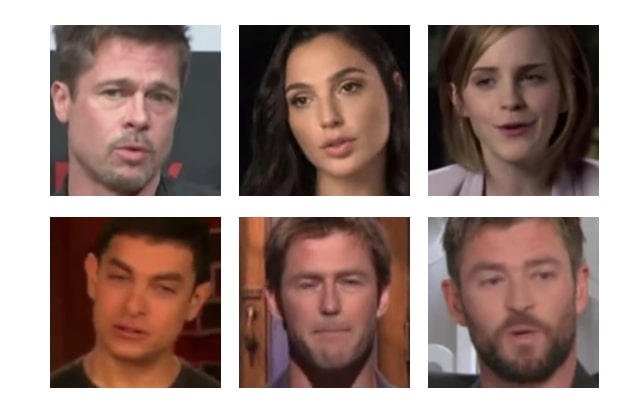} }}%
    \hfill
    \subfloat[AIF~\cite{aif}]{{\includegraphics[width=0.19\textwidth]{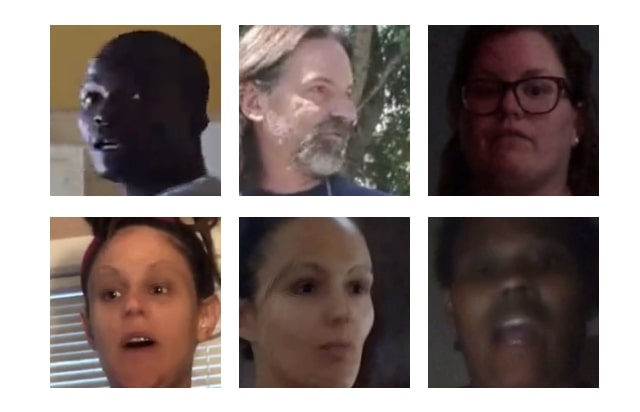} }}%
    
    \caption{Sample frames from the datasets used for evaluation of our experiments. The top row shows the frames from the real videos and bottom row shows the frames from the corresponding fake videos for paired datasets (FF++ and Google DFD) and randomly selected videos for unpaired datasets (Dessa, AIF, and Celeb DF). For FF++, the frames from DF manipulation method are shown.}%
    \label{fig:dataset_snapshot}%
\end{figure}

With the rapid progress in synthetic media generation, a number of datasets focusing on facial forgery detection have been developed recently~\cite{ff_dataset, jiang2020deeperforensics10, Celeb_DF_cvpr20, dfdc_google, korshunov2018deepfakes, dolhansky2019deepfake}.
Since our goal is to generalize across datasets as well as methods, we focus on those with sufficient diversity; see Fig.~\ref{fig:dataset_snapshot}.

\textbf{FaceForensics++ (FF++)~\cite{ff_dataset}:} 
This is the one of the largest datasets in terms of variety and manipulations. 
The dataset contains over 1000 different videos (different identities), with each video manipulated by four different forgery methods, including DeepFakes (DF)~\cite{df_github}, FaceSwap (FS)~\cite{fs_github}, Face2Face (F2F)~\cite{face2face}, NeuralTextures (NT)~\cite{thies2019neural}. 
DF are the most circulated videos on the internet while NT produce high-quality re-enactment videos; hence, we decided to focus on these two manipulation methods. 

\textbf{Google DeepFake Detection (Google DFD)~\cite{dfdc_google}:} 
This dataset consists of 3000 deepfake videos with 28 unique actors in different locations.
To avoid data redundancy, we select all videos from a particular location (i.e., 228 fake and 28 real videos).
In order to obtain an equal number of videos per class, we randomly pick one deepfake video per identity.

\textbf{Celeb DF (v2)~\cite{Celeb_DF_cvpr20}:}
Celeb-DF (v2) is a very challenging dataset of real and synthesized deepfake videos of celebrities with similar visual quality with the online circulated deepfakes. 
It includes 590 original videos and 5639 corresponding DeepFake videos. 
Similar to Google DFD, we randomly pick 590 deepfake videos and use all 590 real videos for a fair comparison.

\textbf{Dessa In-the-Wild~\cite{dessa_dfdc}:} 
The dataset consists of real and fake videos (84 each) of celebrities and politicians created with very high-quality deepfake generation methods.

\textbf{AIF In-the-Wild~\cite{aif}:} 
The AI Foundation (AIF) dataset is a new video dataset which we introduce\footnote{The AIF dataset is donated by the AI Foundation to the authors}. It consists of $\sim$110 unpaired real and deepfake videos.
This is a relatively difficult dataset, with videos captured in low-quality lighting conditions and extreme blurring in some cases; however, this makes it particular interesting as a real-world test case.

\section{Results}\label{results}
\subsection{Experimental Settings}

All of our experiments are run on an Nvidia GeForce GTX 1080 TI and all models use an ILSVRC 2012-pretrained ResNet-18~\cite{He_2016} backbone. 
Our models are implemented in Pytorch trained with an Adam optimizer (default parameters) using a batch-size of 128.
Given that we want to study the transfer results of different loss formulations, we did not focus on hyper-parameter optimization of individual models.

\begin{figure}[htpb]
  \centering
  \includegraphics[width=.75\textwidth]{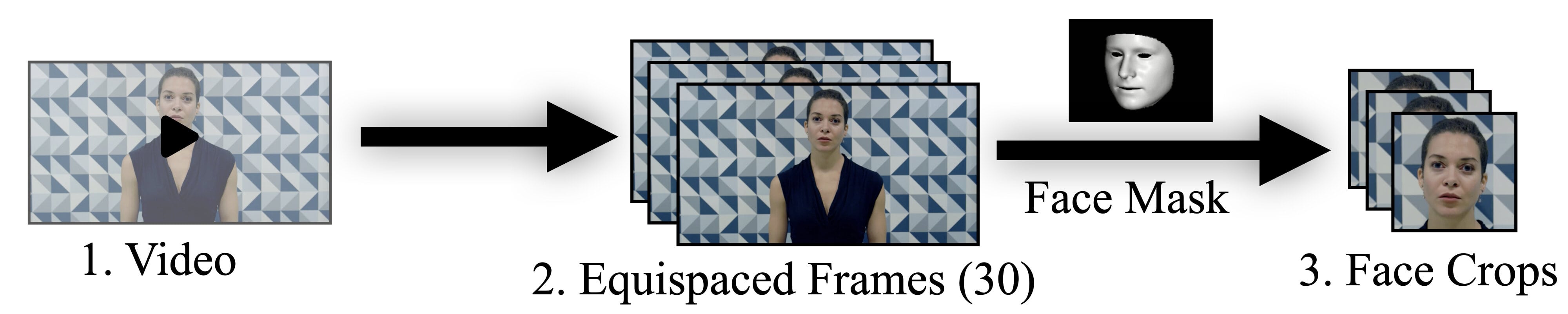}
  \caption{Video pre-processing: We extract 30 frames from every video spaced at equal intervals and automatically crop each frame  ($256 \times 256$ pixels). We ensure that at least $90\%$ of the crop is covered by a face region. FF++ and Google DFD, already provide face masks; for the others, we use OpenCV DLib.}  
  \label{fig:pre_process}
\end{figure}

All videos are converted to individual frames as described in Figure~\ref{fig:pre_process}, and we report accuracies on a per frame basis.
Unless otherwise stated, we apply our proposed spatial mixup augmentation followed by standard flipping to all baselines as well as our method; for fine-tuning it is applied with source images corresponding to the same class.

\subsection{Transfer between Manipulation Methods}
\begin{figure}[htpb]
    \centering
    \subfloat{{\includegraphics[width=0.49\textwidth]{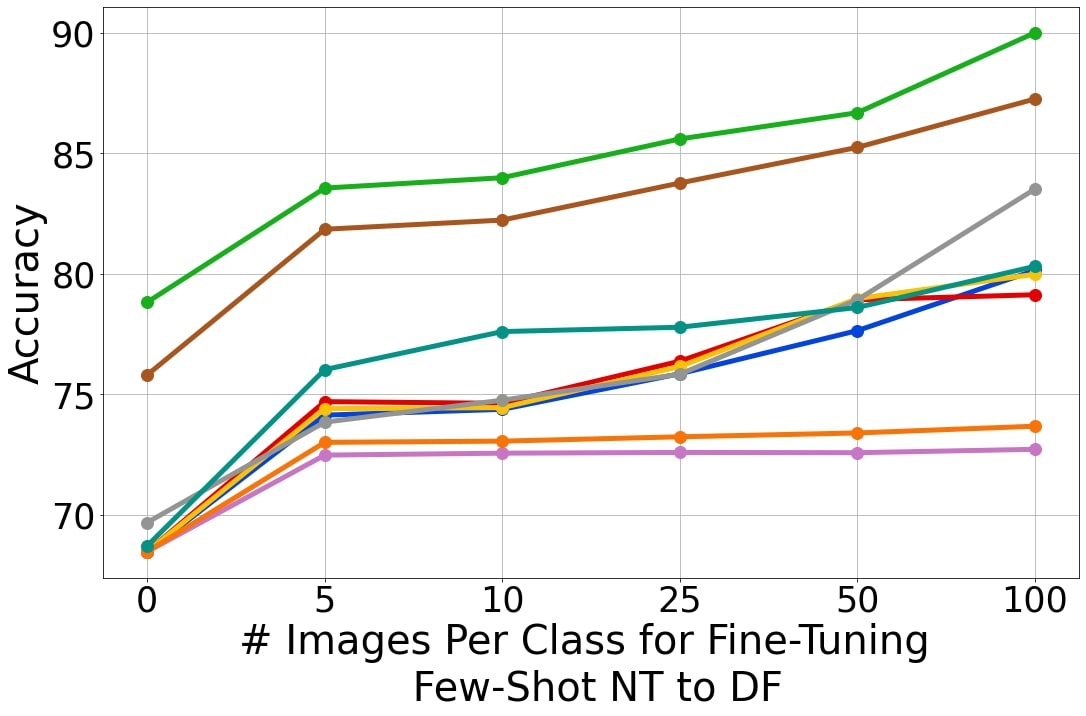} }}%
    \hspace{\fill}
    \subfloat{{\includegraphics[width=0.49\textwidth]{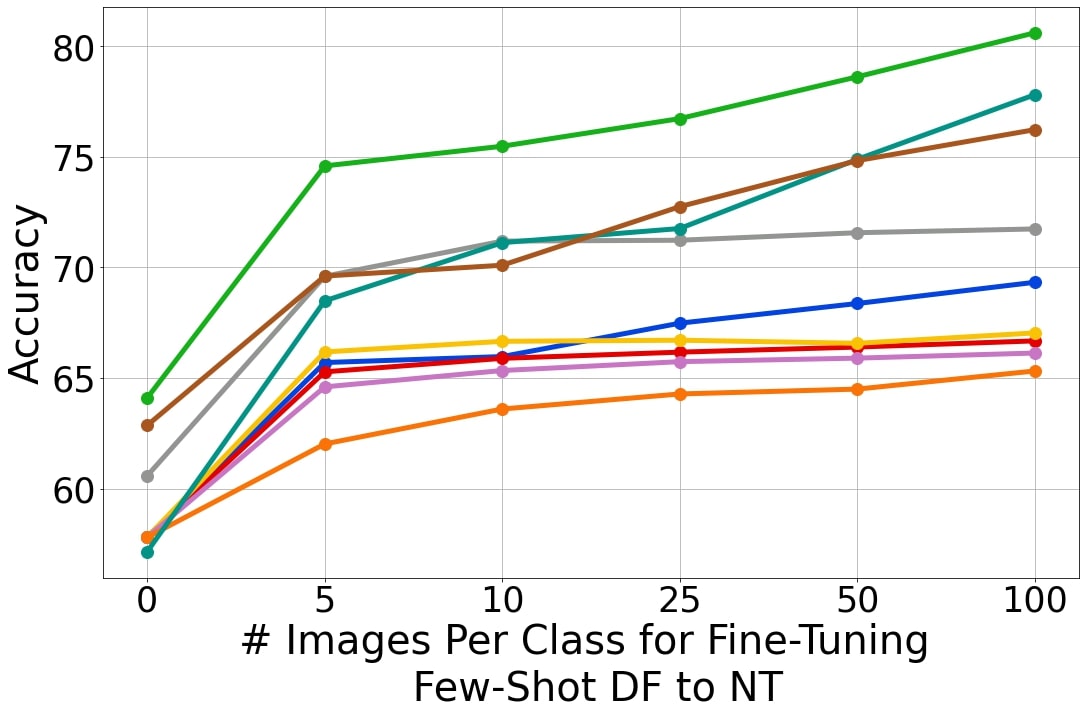} }}%
    \hspace{\fill}
    {\includegraphics[width=1.0\textwidth]{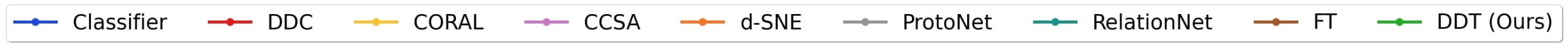}}%
    \caption{Few-shot manipulation transfer\protect\footnotemark: we pre-train a model with one manipulation method and fine-tune with a varying number of images of another manipulation approach. We outperform all the other methods, and achieve 90.01$\%$ accuracy for DF and 80.61$\%$ for NT when only using 100 images.}
    \label{fig:few_shot_ff}
\end{figure}

We investigate how well our method transfers between different manipulation methods, and compare against seven state-of-the-art domain-adaption/few-shot learning methods on the FF++ dataset~\cite{ff_dataset}:
Deep Domain Confusion (DDC)~\cite{tzeng2014deep}, Deep Correlation Alignment (CORAL)~\cite{deep_coral}, Classification and Contrastive Semantic Alignment (CCSA)~\cite{ccsa}, d-SNE~\cite{dsne_2019}, ForensicTransfer (FT)~\cite{forensictransfer}, ProtoNets~\cite{snell2017prototypical}, and RelationNets~\cite{relation_net_2017}. 
The domain adaptation methods~\cite{tzeng2014deep,deep_coral,ccsa,dsne_2019} learn a common latent space for source and target domain and train a classifier over this joint embedding. 
Thus, the zero-shot scenario for these methods boils down to training a classifier on the source domain.
Results are shown in Tab.~\ref{tab:results_zero_shot_ff} and Fig.~\ref{fig:few_shot_ff}.

\begin{table}[htpb]
    \caption{Classification accuracy for zero-shot transfer: the models are trained on one forgery method and then tested on data from a another (unseen) manipulation technique. We achieve comparable results within the same domain; however, our method generalizes significantly better.\\}
      \label{tab:results_zero_shot_ff}
      \centering
  \begin{tabular}{c cx{2cm} cx{2cm} cx{2cm} cx{2cm} c}
  \toprule
 {} & \multicolumn{2}{l}{\textbf{Trained with DF}} & \multicolumn{2}{l}{\textbf{Trained with NT}} & {} \\ 
 {\textbf{Model}} & {DF} & {NT} & {DF} & {NT} & {\textbf{Mean}}\\
\toprule
     {Classifier}& {95.92} & {57.80} & {68.75} & {91.30} & {78.44}\\
     {Prototypical Nets~\cite{snell2017prototypical}} & {\textbf{98.15}} & {60.58} & {69.57} & {\textbf{95.27}} & {80.89} \\
     {Relation Nets~\cite{relation_net_2017}}  & {98.11} & {57.15} & {68.70} & {91.13} & {78.77}\\
     {FT~\cite{forensictransfer}} & {91.29} & {62.86} & {75.50} & {83.50} & {78.28} \\
     {\OURSABB}  & {98.01} & {\textbf{64.10}} & {\textbf{78.82}} & {92.05} & {\textbf{83.25}}\\
    \toprule
  \end{tabular}
  
\end{table}

\newpage

\footnotetext{All the few-shot experiments are averaged over 10 runs.}

\subsection{Transfer between Forgery Datasets}
\begin{figure}[htpb]
    \centering
    \subfloat{{\includegraphics[width=0.31\textwidth]{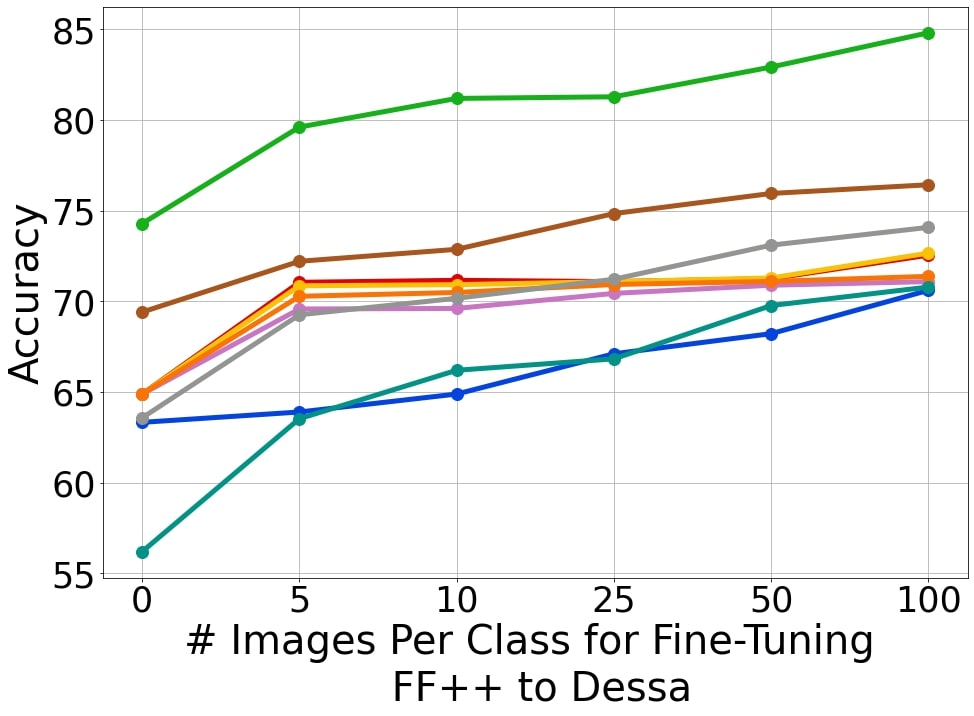} }}%
    \hspace{\fill}
    \subfloat{{\includegraphics[width=0.31\textwidth]{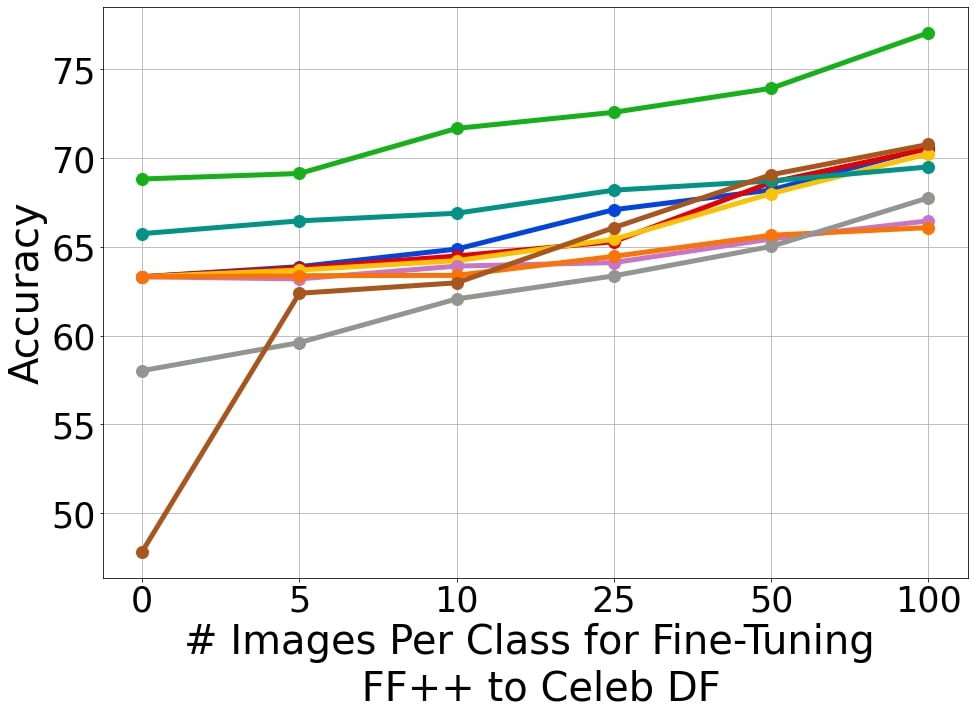} }}%
    \hspace{\fill}
    \subfloat{{\includegraphics[width=0.32\textwidth]{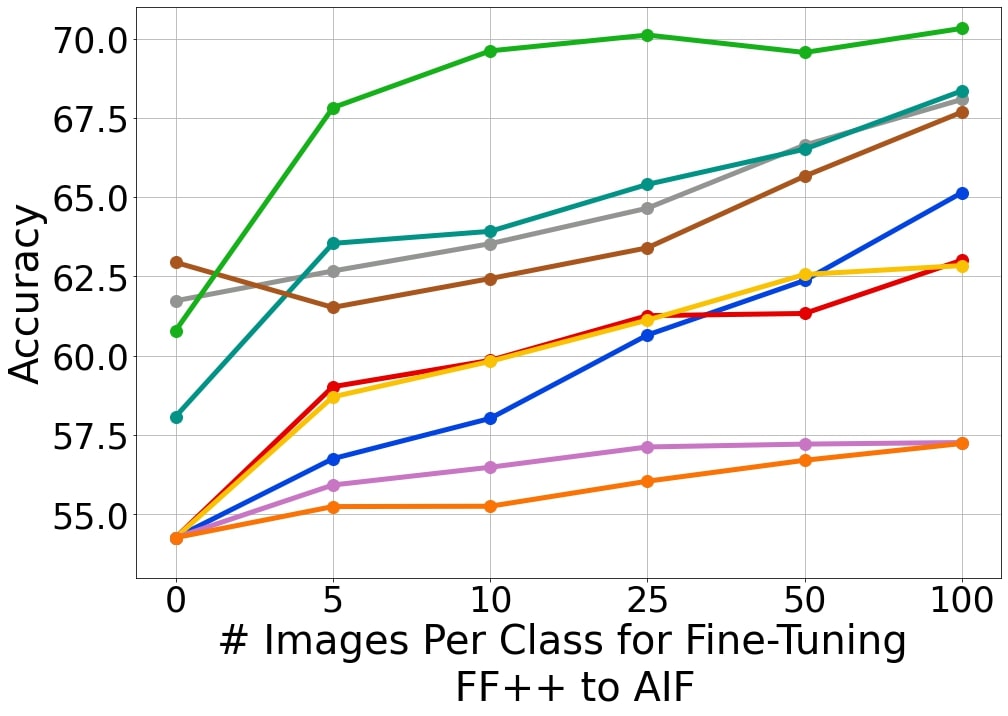} }}%
    \hspace{\fill}
    {\includegraphics[width=1.0\textwidth]{figures_jpg/results/other_datasets/few_shot_new/legends_flat.jpg}}%
     \caption{Few-shot transfer from FF++~\cite{ff_dataset} to three other dataset. Our method outperforms all the other approaches giving 84.80$\%$, 77.07$\%$, 70.32$\%$ accuracy for Dessa, Celeb DF, and AIF dataset, respectively after fine-tuning with 100 images.}%
    \label{fig:few_shot_others}%
\end{figure}

Following the previous section, we observe that NT and DF manipulations exhibit transfer. 
We further validate the claim by exploring the transfer on different datasets. 
Since there are no benchmark datasets to evaluate NT manipulation, we evaluate on a variety of DF manipulation datasets: Google DFD~\cite{dfdc_google}, Celeb DF~\cite{Celeb_DF_cvpr20}, Dessa~\cite{dessa_dfdc}, and AIF~\cite{aif}.
Results for zero-shot transfer from FF++ to other datasets are shown in Tab.~\ref{tab:zero_shot_other_datasets}; few-shot results are visualized in Fig.~\ref{fig:few_shot_others}.

\begin{table}[htpb]
\caption{Zero-shot classification accuracy from FF++ to four other datasets: although there is a significant shift in domains (e.g., Google DFDC and Dessa contain mostly frontal faces in contrast to AIF which varies much more in pose and lightning), our method achieves  transfer performance significantly higher than all baselines (last column).\\}
  \label{tab:zero_shot_other_datasets}
  \centering
  \begin{tabular}{cccccccc}
  \toprule
 {\textbf{Method}} & {\textbf{FF++}}  & {\textbf{Google DFD}} & {\textbf{AIF}} & {\textbf{Dessa}} & {\textbf{Celeb DF}} & {\textbf{Mean}}\\
\toprule
     {Classifier} & {91.02} & {79.87} & {54.26} & {63.45} & {65.32} &  {70.78}\\
     {Prototypical Nets~\cite{snell2017prototypical}} & {\textbf{97.16}} & {71.34} & {61.73} & {63.57} & {58.03} &  {70.36} \\
     {Relation Nets~\cite{relation_net_2017}}  & {96.46} & {72.75} & {58.08} & {56.19} & {63.19} &  {69.84}\\
     {FT~\cite{forensictransfer}} & {84.71} & {67.82} & {\textbf{62.94}} & {69.40} & {47.83}& {66.54} \\
     {\OURSABB}  & {92.23} & {\textbf{81.21}} & {60.79} & {\textbf{74.28}} & {\textbf{68.83}} &  {\textbf{75.47}}\\
    \toprule
  \end{tabular}
  
\end{table}

\subsection{Effect of Spatial Augmentation}
\begin{figure}[htpb]
    \centering
    \subfloat{{\includegraphics[width=0.49\textwidth]{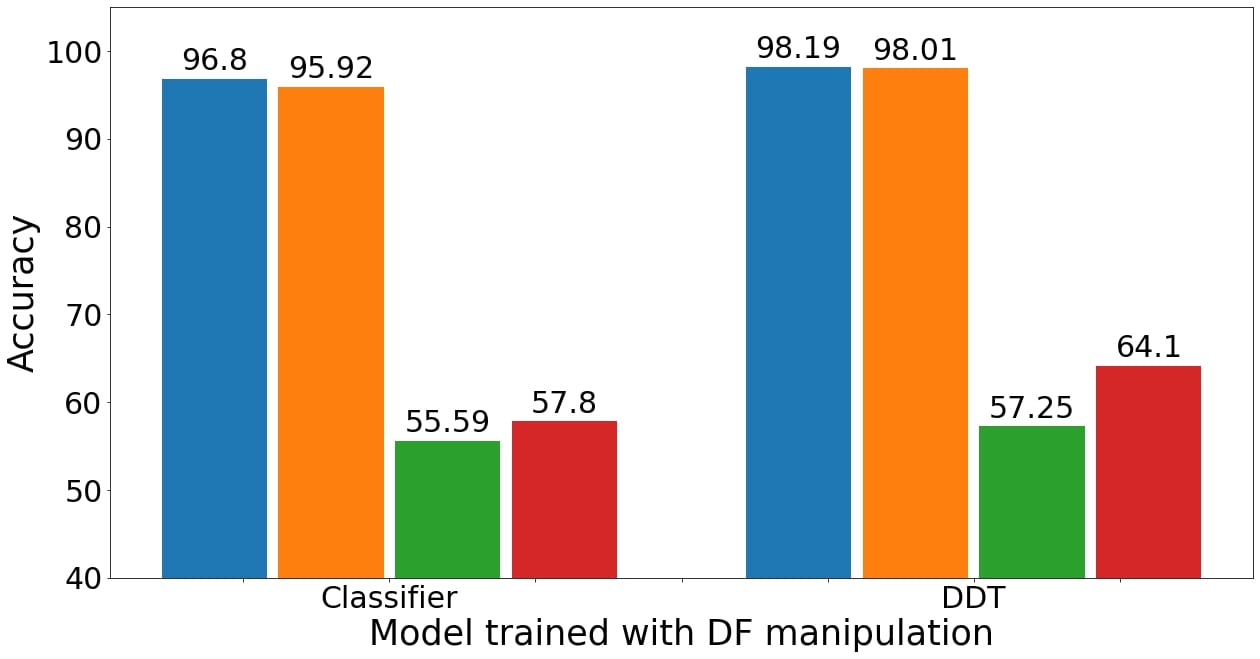} }}%
    \hspace{\fill}
    \subfloat{{\includegraphics[width=0.49\textwidth]{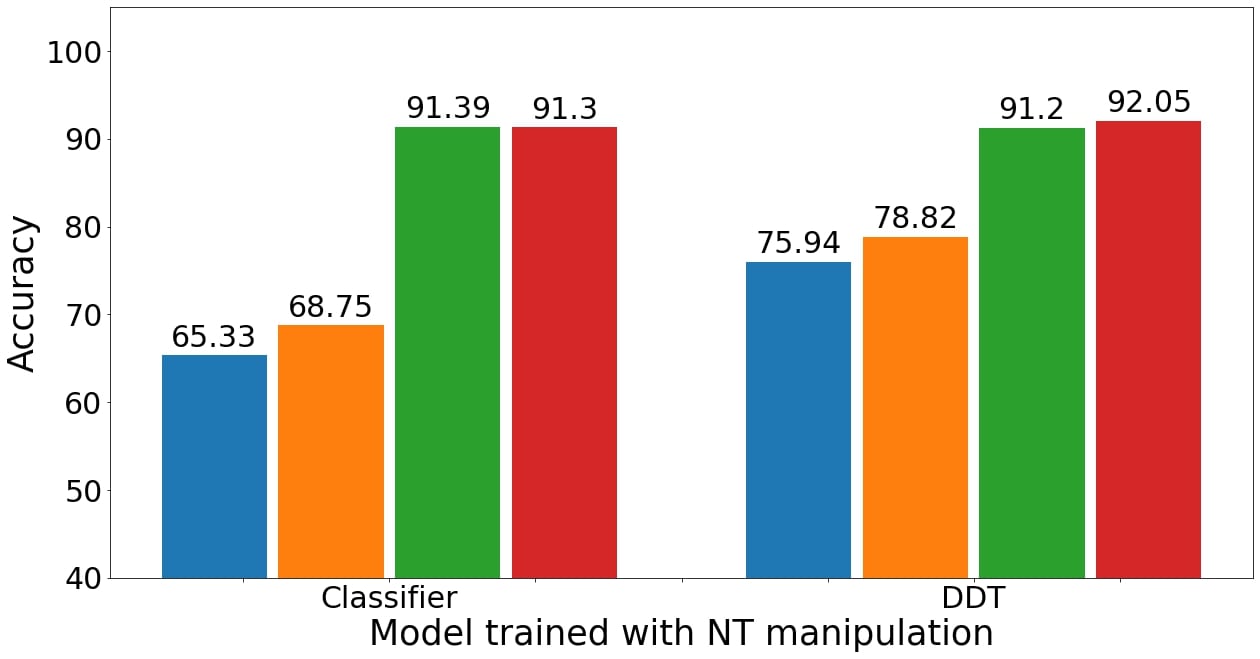} }}%
    \hspace{\fill}
    {\includegraphics[width=.70\textwidth]{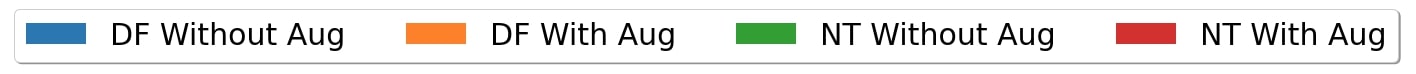}}%
    \caption{Effect of our proposed spatial augmentation for manipulation transfer: applying augmentation during training increases the zero-shot accuracy for both the methods and both experiments: (1) NT to DF ($75.94\%$ to $78.82\%$ for \OURSABBV{}) (2) DF to NT ($57.25\%$ to $64.10\%$ for \OURSABBV{})).
}%
    \label{fig:zero_shot_ff_aug}%
\end{figure}

\begin{figure}[htpb]
    \centering
    \subfloat{{\includegraphics[width=0.23\textwidth]{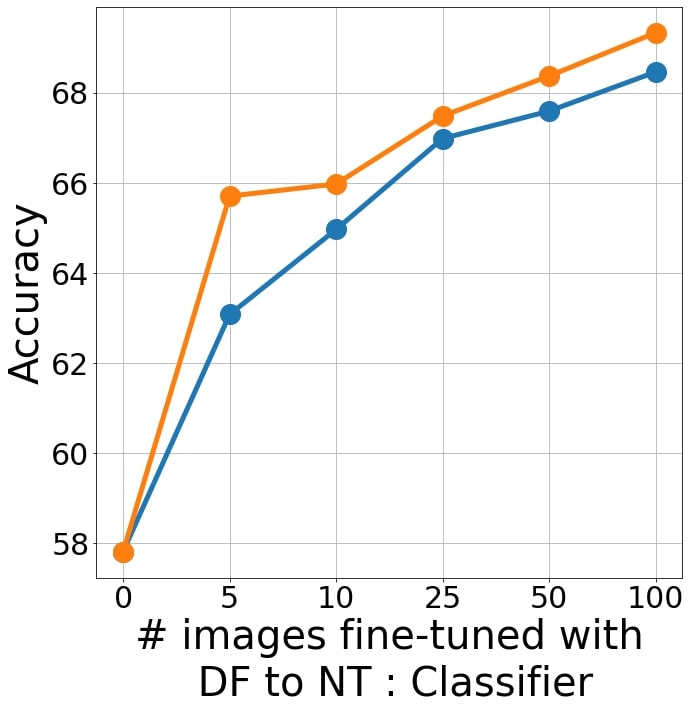} }}%
    \hspace{\fill}
    \subfloat{{\includegraphics[width=0.23\textwidth]{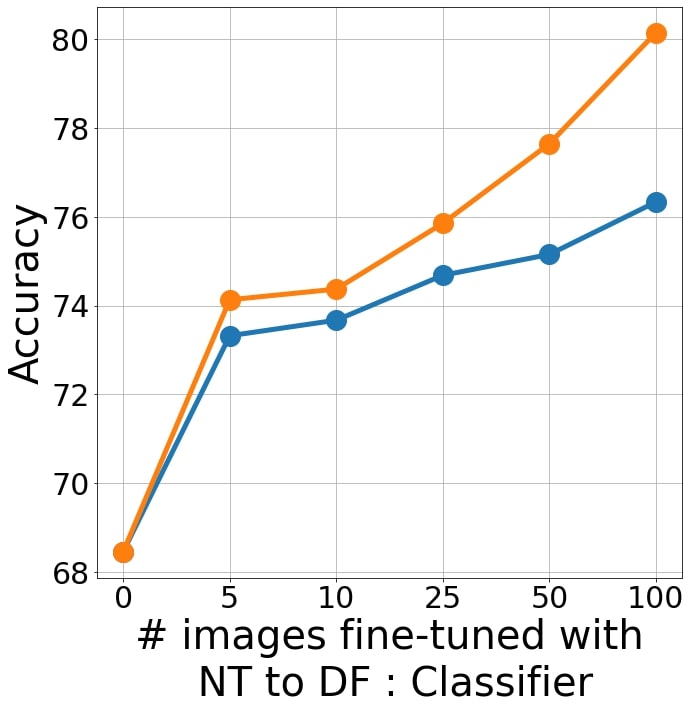} }}%
    \hspace{\fill}
    \subfloat{{\includegraphics[width=0.24\textwidth]{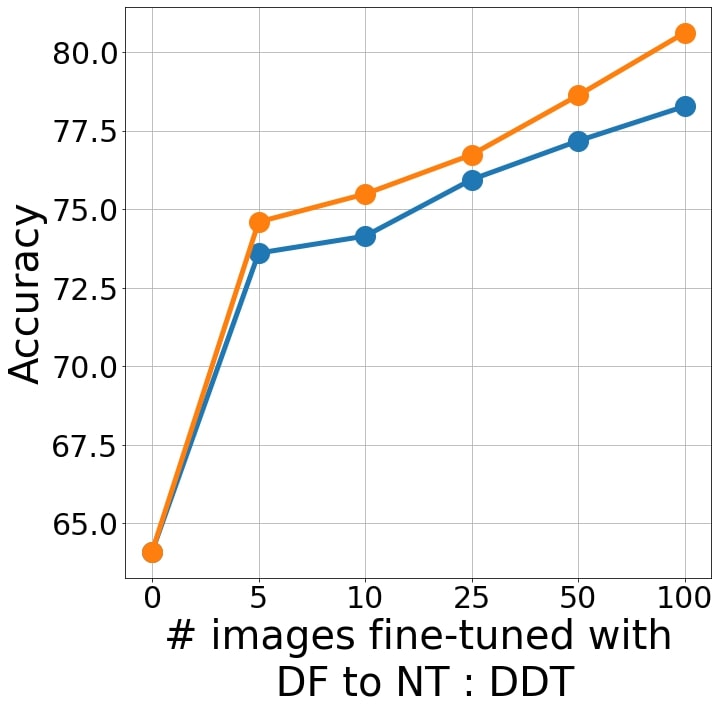} }}%
    \hspace{\fill}
    \subfloat{{\includegraphics[width=0.23\textwidth]{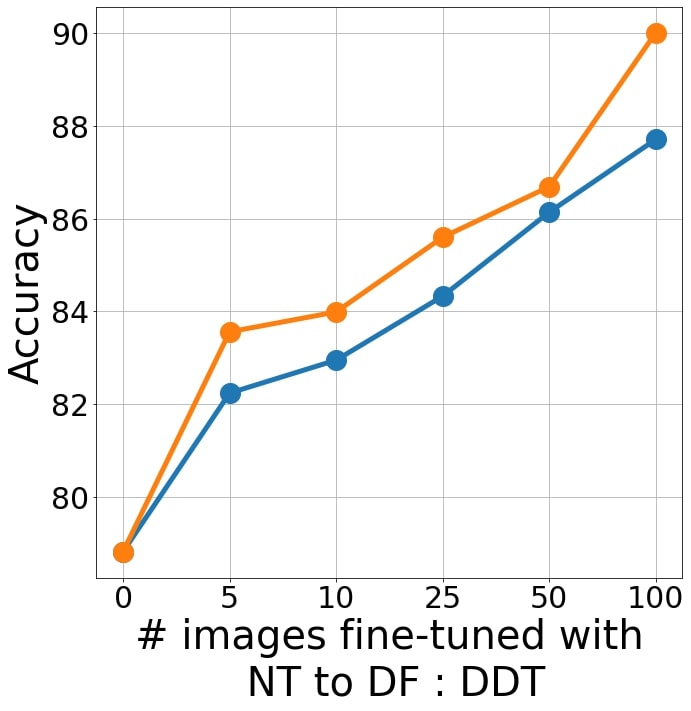} }}%
    \hspace{\fill}
    {\includegraphics[width=.25\textwidth]{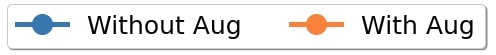}}%
    \caption{Effect of applying spatial augmentation during few-shot manipulation transfer with a binary classifier and our method. Blue and orange colors indicate fine-tuning w/o and w/ spatial augmentation, respectively. We observe a small improvement during fine-tuning across all experiments.
}%
    \label{fig:few_shot_ff_aug}%
\end{figure}

All previous results, including baselines, use our proposed spatial augmentation.
We now evaluate the effectiveness of the data augmentation strategy itself for the binary classifier and our method. 
Fig.~\ref{fig:zero_shot_ff_aug} shows that our augmentation yields similar performance when tested on the same manipulation method; however, it shows significantly better generalization results on unseen forgery methods.
Similar to zero-shot experiments, we evaluate the effect of spatial augmentation for few-shot learning; see Fig.~\ref{fig:few_shot_ff_aug}.
To this end, we mix images from the target domain with the images from the source domain. 
Every time we randomly pick a different image of the same class from the source domain to mix with the images of the target domain. 
In contrast to zero-shot self-manipulation results, where we did not see a significant effect on the same manipulation method, we now notice a small improvement using spatial augmentation.
For instance, for the NT to DF manipulation transfer task, after fine-tuning with 100 images accuracy improves by $3.81\%$ ($76.33\%$ to $80.14\%$) for the binary classifier and $2.3\%$ ($87.71\%$ to $90.01\%$) for our approach.

\vspace{-0.3cm}

\vspace{-0.3cm}

\section{Conclusion}
We have presented Deep Distribution Transfer, a new method for generalized zero and few-shot transfer learning for facial forgery detection.
The core idea of our method is a new distribution-based loss formulation that can be efficiently trained to bridge the gap between domains of different facial forgery methods or unseen datasets.
In a series of experiments, we show that the proposed method transfers well between widely-used face forgery data datasets while outperforming state-of-the-art baselines by a significant margin in both zero-shot and few-shot settings.
For instance, we achieve a 4.88\% higher detection accuracy for zero-shot and 8.38\% for the few-shot case transferred from the FaceForensics++ to Dessa dataset.
Overall, we believe that our generalized forensics transfer method is an important stepping stone towards making automated media forensics viable in practical settings, given that we ultimately need to have a reliable detector which is capable of handling previously unseen fakes and new data sources. 

\newpage

\section*{Broader Impact}
The rapid development of facial manipulation methods such as deepfakes has become a severe issue in the context of social media and online video platforms, hence making the research on media forensics a critical research area within the machine learning community.
Although there have been works focusing on automated detection methods with neural networks, such methods train and test within the same domain of data and methods.
In this work, we directly address this limitation, and aim to generalize across forgery techniques and datasets with a new zero-shot and few-shot formulation.
We believe that this transfer is critical towards making learned forgery detection methods practical, given that new fake methods appear at a rapid rate and it is unrealistic to expect to have access to a large training corpus for each new method.
Ultimately, we hope that our method is a first stepping stone towards automated forgery detection on in-the-wild videos, for instance on Youtube, Facebook, or Twitter, rather than focusing on individual (potentially biased) academic datasets.
In addition, we hope that the release of our code and trained models can be already of practical use to fact checkers as well as other researchers building upon on our work.

\section*{Acknowledgements}
We would like to thank Andreas R\"ossler for helpful discussions and Davide Cozzolino for providing the Forensic Transfer~\cite{forensictransfer} source code.
We gratefully acknowledge the support of this research by the AI Foundation, a TUM-IAS Rudolf M\"o{\ss}bauer Fellowship, the ERC Starting Grant Scan2CAD (804724), and Google Faculty Award.

{


}

\newpage
\begin{appendix}
\section{Datasets}

\subsection{Dataset Statistics}
We evaluate our zero/few-shot transfer approach on five different forgery detection benchmark and in-the-wild video datasets: FaceForensics++~\cite{ff_dataset}, Google DFD~\cite{dfdc_google}, Celeb DF~\cite{Celeb_DF_cvpr20}, Dessa~\cite{dessa_dfdc}, and AIF~\cite{aif} dataset. 
The exact details of train-test split used for our experiments are listed in Tab.~\ref{tab:dataset_split}.
\begin{table}[htpb]
  \centering
  \caption[]{Dataset statistics showing videos per class. For FF++ and Dessa, we use the already provided split; for the other datasets, we provide our own train-test split. Note that for Google DFD, we have very few real videos; hence, we only evaluate zero-shot performance for this dataset. For all other datasets, we explore both zero-shot and few-shot transfer. All results in the main paper are reported on these train-test splits.}
  \label{tab:dataset_split}
  \begin{tabular}{c c c c c c }
  \toprule
    \textbf{Mode} & \textbf{FF++}~\cite{ff_dataset} & \textbf{Google DFD}~\cite{dfdc_google} & \textbf{Celeb DF}~\cite{Celeb_DF_cvpr20} & \textbf{Dessa}~\cite{dessa_dfdc} & \textbf{AIF}~\cite{aif}  \\
    \toprule
    \textbf{Train} & 720 & - & 500 & 70 & 12  \\
    
    \textbf{Val} & 140 & - & - & - & - \\
    
    \textbf{Test} & 140 & 28 & 90 & 14 & 99  \\
    \toprule
  \end{tabular}
  
\end{table}

\subsection{Detailed of Google DFD Selection Strategy}
\begin{figure}[htpb]
    \centering
    \subfloat[Same actor but different location]{{\includegraphics[width=0.37\textwidth]{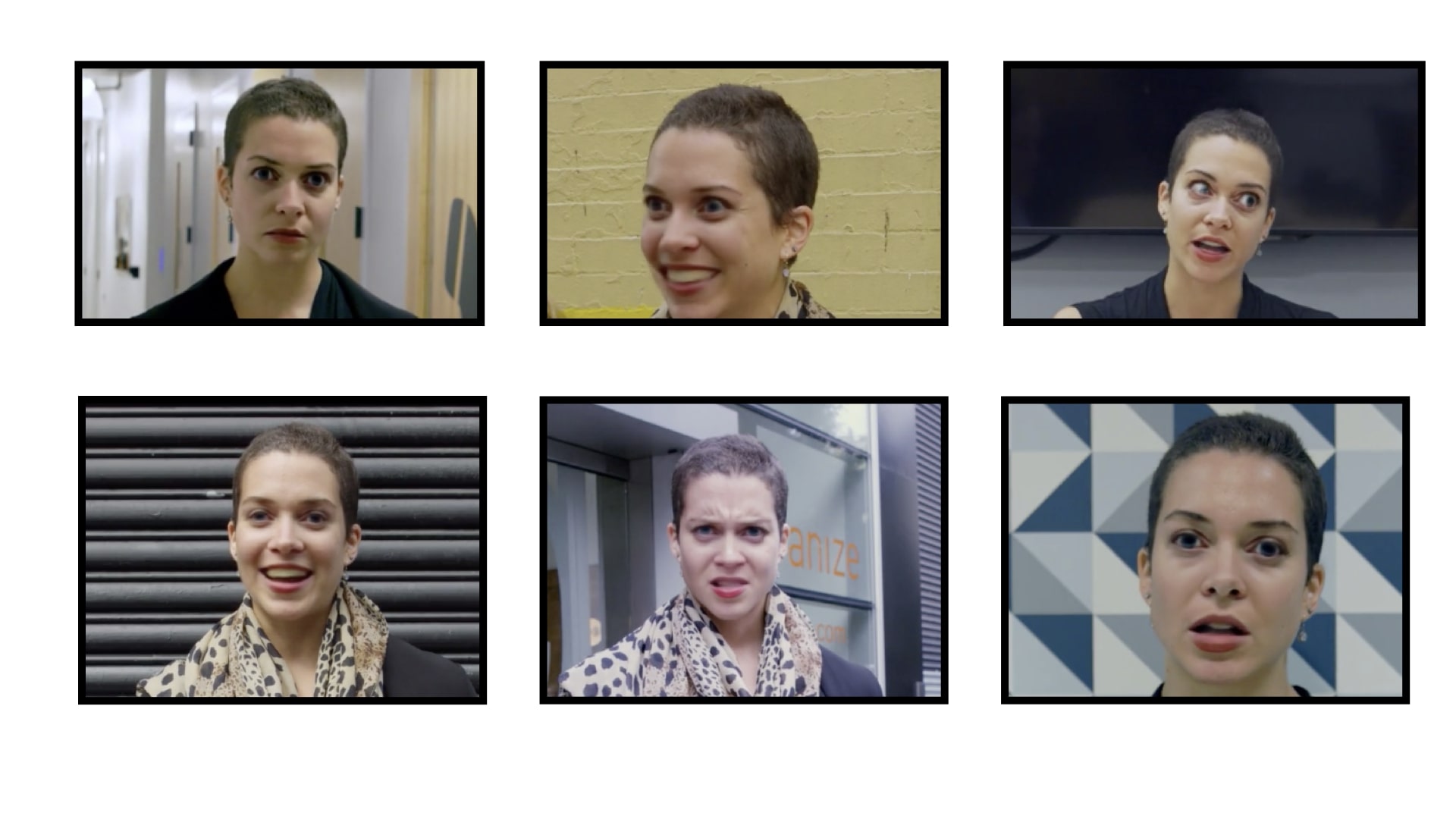} }}%
    \qquad
    \subfloat[Different actors but same location.]{{\includegraphics[width=0.37\textwidth]{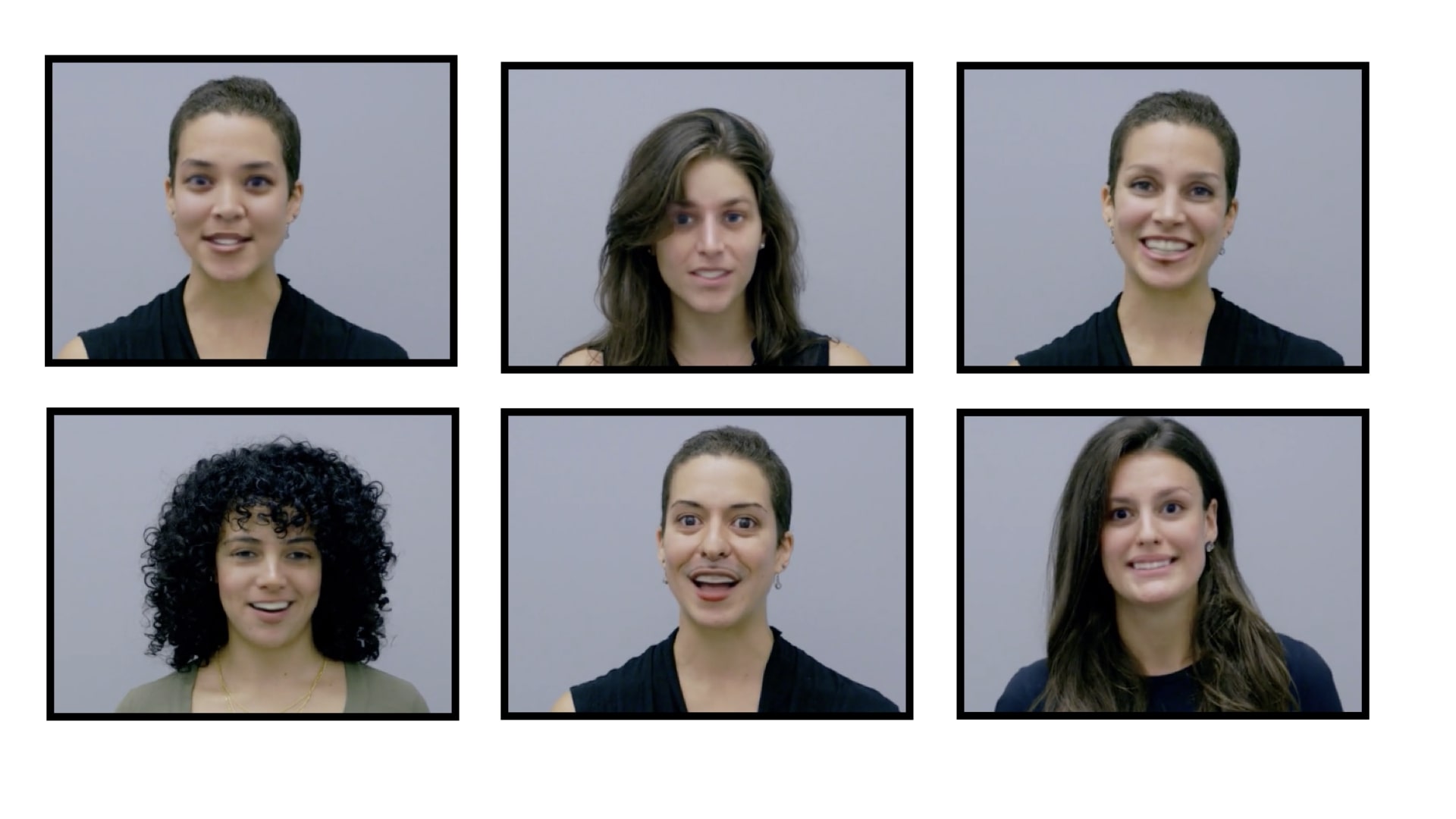} }}%
    \caption{Google DFD dataset selection strategy: (a) we have multiple videos of the same actor (recorded in different locations) with deepfakes are created for all the real videos; i.e., we have multiple videos with the same facial identity in different locations. (b) We pick one location (podium in our case) and use the videos (real and fake) for this particular location.}%
    \label{fig:dfdc_dataset_selection}%
\end{figure}

In order to avoid redundant videos in Google DFD, we select 228 fake and 28 real videos following the strategy described in Fig.~\ref{fig:dfdc_dataset_selection}.

\section{Details on Hyperparameters}
We use the ILSVRC 2012-pretrained ResNet-18~\cite{He_2016} network as a backbone for our experiments to obtain a 256-dimensional embedding vector. 
We then apply one or more dense layers on top of this embedding based on what transfer learning approach we are training with. 
For ForensicTransfer (FT)~\cite{forensictransfer}, we used the original architecture and setup as proposed in the original paper.

For pre-training (zero-shot) experiments, we use a learning rate of 1e-3 with a decay when validation loss plateaus for five consecutive epochs, early stopping with the patience of 10 successive epochs on the validation loss. 
For fine-tuning (few-shot) and domain transfer experiments, we use a learning rate of 1e-5 without any decay, early stopping with the patience of 30 consecutive epochs on the training loss since no validation set is available. 
During fine-tuning, we use the same seed across runs and for each method to ensure consistency. 

\section{Additional Experiments}

\subsection{Effect of Adding More Manipulation Methods}
In this section, we evaluate how well a model trained with DF or NT or both manipulations from FF++ dataset~\cite{ff_dataset} is able to detect fake videos from a different and unseen dataset. 

\begin{table}[htpb]
  \centering
  \caption{Zero-shot transfer comparison from different manipulation methods. A model pre-trained with different manipulation methods from FF++~\cite{ff_dataset} dataset is evaluated on four different datasets: Google DFD, AIF, Dessa, and Celeb DF. We experiment with three combination of manipulations, models trained with DF only, NT only and both manipulations.}
  \label{tab:zero_shot_forgery_other_datasets}
  \begin{tabular}{cccccccc}
  \toprule
 {\textbf{Method}} &  {\textbf{Manipulation}} & {\textbf{Google DFD}} & {\textbf{AIF}} & {\textbf{Dessa}} & {\textbf{Celeb DF}} & {\textbf{Mean}}\\
\toprule
    {} & {DF} & {79.10} & {53.78} & {55.35} & {63.37} &  {62.90}\\
     {Classifier} & {NT} & {68.01} & {53.01} & {58.80} & {52.90} &  {58.18}\\
     {} & {DF+NT} & {\textbf{79.87}} & {\textbf{54.26}} & {\textbf{63.45}} & {\textbf{65.32}} &  {\textbf{65.72}}\\
     \midrule
     {}  & {DF} & {\textbf{81.94}} & {53.06} & {59.28} & {67.98} &  {65.56}\\
     {\OURSABB}  & {NT} & {75.38} & {55.47} & {59.52} & {66.99} &  {64.34}\\
     {}  & {DF+NT} & {81.22} & {\textbf{60.79}} & {\textbf{74.28}} & {\textbf{68.83}} &  {\textbf{71.28}}\\
    \toprule
  \end{tabular}
  
\end{table}

Tab.~\ref{tab:zero_shot_forgery_other_datasets} shows that a model pre-trained with both DF and NT manipulations boosts accuracy by 0.85\%(from 67.98\% to 68.83\%) for Celeb DF, 5.32\% (from 55.47\% to 60.79\%) for AIF and a large 14.76\% (from 59.52\% to 74.28\%) for Dessa dataset, for our proposed method. 
Similarly, for the naive classifier, the accuracy is boosted by 1.95\%(from 63.37\% to 65.32\%) for Celeb DF, 0.48\% (from 53.78\% to 54.26\%) for AIF and 4.65\% (from 58.80\% to 63.45\%) for Dessa dataset. 
An important observation is that adding more variety in manipulations methods for pre-training yields to better generalizability; i.e., we obtain better detection results on different, previously unseen, datasets.
This is very promising and shows a path forward to the future of forgerey detection when more datasets will become publicily available, which would allow us to obtain even stronger models with our method.
At the same time, there is no performance degradation on the Google DFD dataset when we move from DF pre-trained model to DF+NT pretrained model. 
Hence, adding more variety in terms of manipulation methods for pre-training does not come at the cost of already included target forgery techniques.

\subsection{Effect of Applying Spatial Augmentation on Other Datasets}
In this section, we study the effect of applying our proposed augmentation on other datasets.
We evaluate the genralization results on two challenging AIF and Dessa datasets.
\begin{figure}[htpb]
  \centering
   \includegraphics[width=1.0\textwidth]{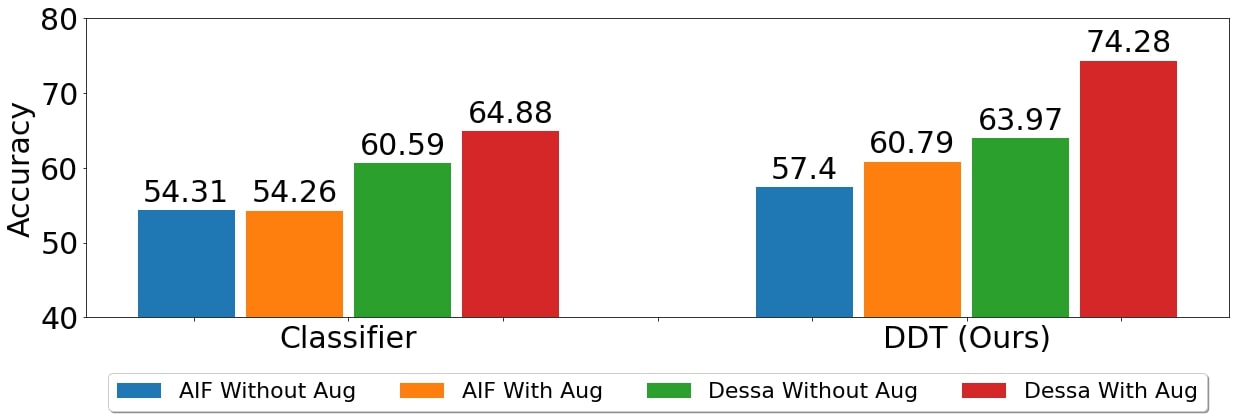}
  \caption[]{Zero-shot transfer comparison from FF++ to other datasets for a naive classifier and our method. All the models are pre-trained on (NT + DF) manipulation both, with and without spatial mixup augmentation.} 
  \label{fig:zero_shot_other_datasets_aug}
 \end{figure}
 
Fig.~\ref{fig:zero_shot_other_datasets_aug} shows that applying spatial mixup augmentation during pre-training generalizes better to other datasets as well. 
For the Dessa dataset, we observe 4.29\% (from 60.59\% to 64.88\%) improvement for Classifier and a massive 10.31\% improvement (from 63.97\% to 74.28\%) for \OURSABBV{}.

\begin{figure}[htpb]
    \centering
    \subfloat{{\includegraphics[width=0.23\textwidth]{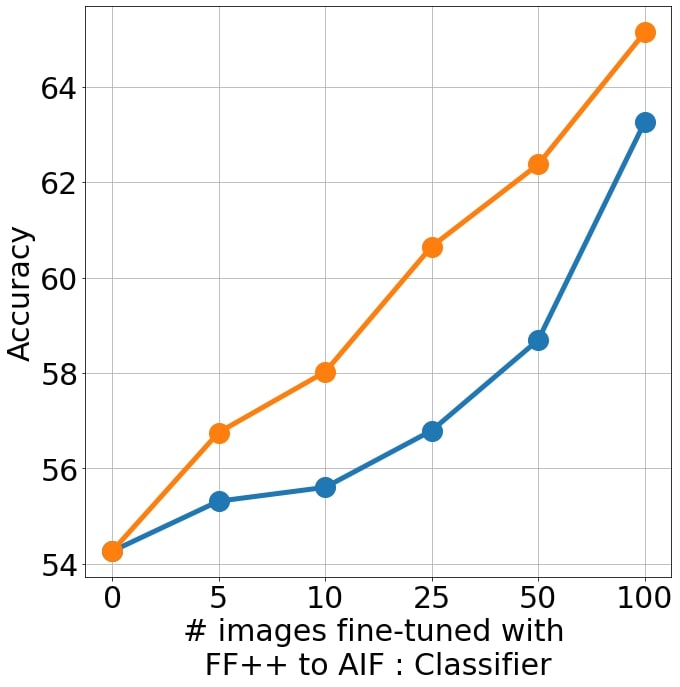} }}%
    \hspace{\fill}
    \subfloat{{\includegraphics[width=0.23\textwidth]{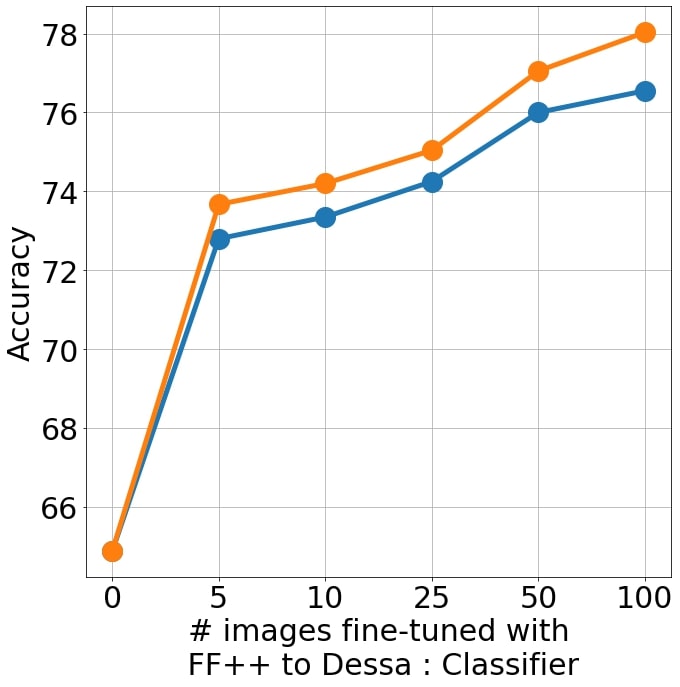} }}%
    \hspace{\fill}
    \subfloat{{\includegraphics[width=0.23\textwidth]{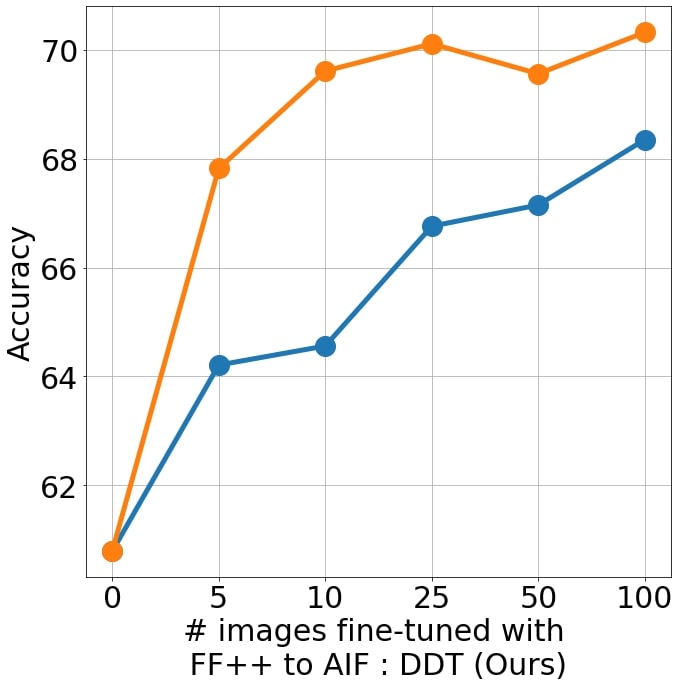} }}%
    \hspace{\fill}
    \subfloat{{\includegraphics[width=0.23\textwidth]{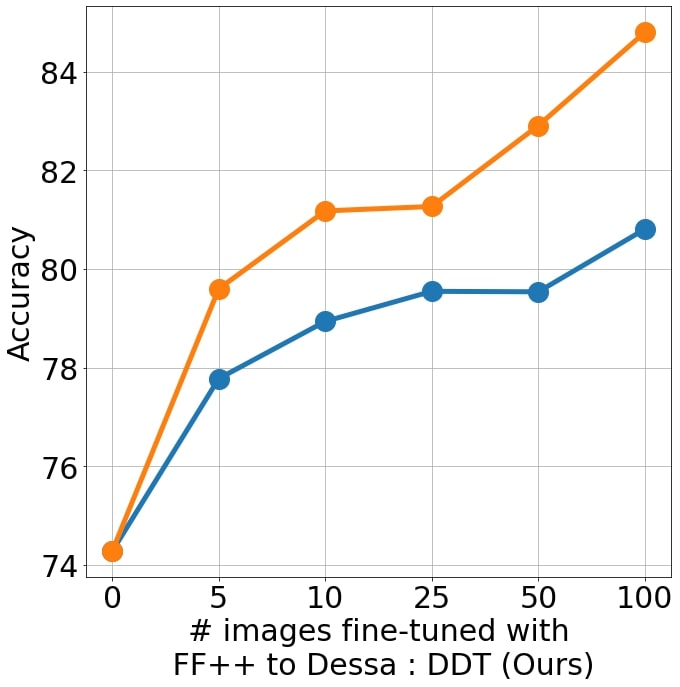} }}%
    \hspace{\fill}
    {\includegraphics[width=.25\textwidth]{figures_jpg/results/faceforensics/few_shot_aug/legends_aug.jpg}}%
    \caption{Effect of applying our spatial augmentation during few-shot transfer to other datasets for a binary classifier and our method. During fine-tuning, we again observe consistent improvements across all experiments.
}%
    \label{fig:few_shot_other_aug}%
\end{figure}

\section{Result Tables}
In this section, we document the line graphs of all the few shot experiments.

\begin{table}[htpb]
    \caption{Few-shot manipulation transfer from NT to DF.}
      \label{tab:results_few_shot_nt_to_df}
      \centering
  \begin{tabular}{c c c c c c c c c c}
  \toprule
    {\textbf{Few-Shot}} & {} & {} & {Deep} & {} & {} & {Prototypical} & {Relation} & {Forensic} & {\textbf{DDT}}\\
    {\textbf{Images}} & {Classifier} & {DDC} & {CORAL} & {CCSA} & {d-SNE} & {Nets} & {Nets} & {Transfer} & {\textbf{(Ours)}}\\
\toprule
     {0 image} & {68.46} & {68.46} & {68.46} & {68.46} & {68.46} & {69.67} & {68.70} & {75.80} & \textbf{78.82} \\
     {5 images} & {74.13} & {74.69} & {74.40} & {72.47} & {73.0} & {73.84} & {76.02} & {81.85} & \textbf{83.56}  \\
     {10 images} & {74.37} & {74.62} & {74.44} & {72.55} & {73.05} & {74.75} & {77.60} & {82.23} & \textbf{83.99} \\
     {25 images} & {75.85} & {76.37} & {76.15} & {72.58} & {73.23} & {75.83} & {77.78} & {83.77} & \textbf{85.60} \\
     {50 images} & {77.64} & {78.92} & {78.97} & {72.57} & {73.39} & {78.92} & {78.60} & {85.25} & \textbf{86.69} \\
     {100 images} & {80.14} & {79.13} & {79.98} & {72.71} & {73.67} & {83.52} & {80.31} & {87.26} & \textbf{90.01} \\
    \toprule
  \end{tabular}
  
\end{table}

\begin{table}[htpb]
    \caption{Few-shot manipulation transfer from DF to NT.}
      \label{tab:results_few_shot_df_to_nt}
      \centering
  \begin{tabular}{c c c c c c c c c c}
  \toprule
    {\textbf{Few-Shot}} & {} & {} & {Deep} & {} & {} & {Prototypical} & {Relation} & {Forensic} & {\textbf{DDT}}\\
    {\textbf{Images}} & {Classifier} & {DDC} & {CORAL} & {CCSA} & {d-SNE} & {Nets} & {Nets} & {Transfer} & {\textbf{(Ours)}}\\
\toprule
     {0 image} & {57.80} & {57.80} & {57.80} & {57.80} & {57.80} & {60.58} & {57.15} & {62.86} & \textbf{64.10} \\
     {5 images} & {65.70} & {65.28} & {66.18} & {64.60} & {62.02} & {69.61} & {68.49} & {69.61} & \textbf{74.59}  \\
     {10 images} & {65.97} & {65.89} & {66.66} & {65.34} & {63.60} & {71.19} & {71.12} & {70.10} & \textbf{75.48} \\
     {25 images} & {67.48} & {66.17} & {66.71} & {65.74} & {64.28} & {71.23} & {71.76} & {72.75} & \textbf{76.73} \\
     {50 images} & {68.37} & {66.40} & {66.57} & {65.90} & {64.50} & {71.57} & {74.89} & {74.83} & \textbf{78.62} \\
     {100 images} & {69.33} & {66.68} & {67.04} & {66.13} & {65.32} & {71.74} & {77.81} & {76.23} & \textbf{80.61} \\
    \toprule
  \end{tabular}
  
\end{table}

\begin{table}[htpb]
    \caption{Few-shot transfer from FF++ to Dessa\\}
    \vspace{0.1cm}
      \label{tab:results_few_shot_ff_to_dessa}
      \centering
  \begin{tabular}{c c c c c c c c c c}
  \toprule
    {\textbf{Few-Shot}} & {} & {} & {Deep} & {} & {} & {Prototypical} & {Relation} & {Forensic} & {\textbf{DDT}}\\
    {\textbf{Images}} & {Classifier} & {DDC} & {CORAL} & {CCSA} & {d-SNE} & {Nets} & {Nets} & {Transfer} & {\textbf{(Ours)}}\\
\toprule
     {0 image} & {64.88} & {64.88} & {64.88} & {64.88} & {64.88} & {63.57} & {56.19} & {69.40} & \textbf{74.28} \\
     {5 images} & {73.67} & {71.04} & {70.85} & {69.57} & {70.27} & {69.25} & {63.52} & {72.20} & \textbf{79.60}  \\
     {10 images} & {74.20} & {71.15} & {70.91} & {69.60} & {70.48} & {70.16} & {66.19} & {72.86} & \textbf{81.18} \\
     {25 images} & {75.04} & {71.08} & {71.08} & {70.43} & {70.92} & {71.21} & {66.82} & {74.83} & \textbf{81.27} \\
     {50 images} & {77.05} & {71.24} & {71.28} & {70.88} & {71.09} & {73.09} & {69.76} & {75.94} & \textbf{82.91} \\
     {100 images} & {78.03} & {72.54} & {72.65} & {71.08} & {71.37} & {74.07} & {70.78} & {76.42} & \textbf{84.80} \\
    \toprule
  \end{tabular}
\end{table}

\begin{table}[htpb]
    \caption{Few-shot transfer from FF++ to Celeb DF\\}
    \vspace{0.1cm}
      \label{tab:results_few_shot_ff_to_celeb}
      \centering
  \begin{tabular}{c c c c c c c c c c}
  \toprule
    {\textbf{Few-Shot}} & {} & {} & {Deep} & {} & {} & {Prototypical} & {Relation} & {Forensic} & {\textbf{DDT}}\\
    {\textbf{Images}} & {Classifier} & {DDC} & {CORAL} & {CCSA} & {d-SNE} & {Nets} & {Nets} & {Transfer} & {\textbf{(Ours)}}\\
\toprule
     {0 image} & {63.32} & {63.32} & {63.32} & {63.32} & {63.32} & {58.03} & {65.75} & {47.83} & \textbf{68.83} \\
     {5 images} & {63.89} & {63.82} & {63.70} & {63.20} & {63.38} & {59.61} & {66.47} & {62.39} & \textbf{69.14}  \\
     {10 images} & {64.88} & {64.49} & {64.23} & {63.92} & {63.40} & {62.07} & {66.90} & {62.98} & \textbf{71.68} \\
     {25 images} & {67.10} & {65.30} & {65.43} & {64.12} & {64.47} & {63.37} & {68.20} & {66.09} & \textbf{72.59} \\
     {50 images} & {68.21} & {68.64} & {68.01} & {65.45} & {65.66} & {65.03} & {68.72} & {69.07} & \textbf{73.94} \\
     {100 images} & {70.60} & {70.54} & {70.25} & {66.47} & {66.09} & {67.77} & {69.51} & {70.78} & \textbf{77.07} \\
    \toprule
  \end{tabular}
\end{table}

\begin{table}[htpb]
    \caption{Few-shot transfer from FF++ to AIF \\}
    \vspace{0.1cm}
      \label{tab:results_few_shot_ff_to_aif}
      \centering
  \begin{tabular}{c c c c c c c c c c}
  \toprule
    {\textbf{Few-Shot}} & {} & {} & {Deep} & {} & {} & {Prototypical} & {Relation} & {Forensic} & {\textbf{DDT}}\\
    {\textbf{Images}} & {Classifier} & {DDC} & {CORAL} & {CCSA} & {d-SNE} & {Nets} & {Nets} & {Transfer} & {\textbf{(Ours)}}\\
\toprule
     {0 image} & {54.26} & {54.26} & {54.26} & {54.26} & {54.26} & {61.73} & {58.08} & \textbf{62.94} & 60.79 \\
     {5 images} & {56.75} & {59.02} & {58.70} & {55.92} & {55.24} & {62.67} & {63.54} & {61.52} & \textbf{67.82}  \\
     {10 images} & {58.02} & {59.85} & {59.82} & {56.48} & {55.25} & {63.53} & {63.92} & {62.43} & \textbf{69.61} \\
     {25 images} & {60.65} & {61.26} & {61.12} & {57.12} & {56.04} & {64.65} & {65.40} & {63.40} & \textbf{70.11} \\
     {50 images} & {62.37} & {61.33} & {62.56} & {57.21} & {56.70} & {66.64} & {66.51} & {65.66} & \textbf{69.56} \\
     {100 images} & {65.14} & {63.011} & {62.84} & {57.26} & {57.23} & {68.08} & {68.35} & {67.68} & \textbf{70.32} \\
    \toprule
  \end{tabular}
\end{table}

\begin{table}[htpb]
      \centering
      \caption{Few Shot Transfer with and without spatial augmentation for DF to NT within the FF++ dataset.}
      \label{tab:results_few_shot_aug_df_to_nt}
  \begin{tabular}{c| c c| c c}
  \toprule
 {\textbf{Few-Shot}} & \multicolumn{2}{c|}{\textbf{Classifier}} & \multicolumn{2}{c}{\textbf{DDT (Ours)}} \\ 
 {\textbf{Images}} & {Without Aug} & {With Aug} & {Without Aug} & {With Aug} \\
\toprule
     {0 image}& {57.80} & {57.80} & {64.10} & {64.10}\\
     {5 images}& {63.07} & {65.70} & {73.60} & {74.59} \\
     {10 images}& {64.96} & {65.97} & {74.14} & {75.48} \\
     {25 image}& {66.98} & {67.48} & {75.94} & {76.73} \\
     {50 image}& {67.59} & {68.37} & {77.17} & {78.62} \\
     {100 image}& {68.46} & {69.33} & {78.27} & {80.61} \\
    
    \toprule
  \end{tabular}
   
\end{table}

\begin{table}[htpb]
      \centering
      \caption{Few-shot transfer with and without spatial augmentation for NT to DF within the FF++ dataset.}
      \label{tab:results_few_shot_aug_nt_to_df}
  \begin{tabular}{c| c c| c c}
  \toprule
 {\textbf{Few-Shot}} & \multicolumn{2}{c|}{\textbf{Classifier}} & \multicolumn{2}{c}{\textbf{DDT (Ours)}} \\ 
 {\textbf{Images}} & {Without Aug} & {With Aug} & {Without Aug} & {With Aug} \\
\toprule
     {0 image}& {68.46} & {68.46} & {78.82} & {78.82} \\
     {5 images}& {73.32} & {74.13} & {82.24} & {83.56} \\
     {10 images}& {73.67} & {74.37} & {82.95} & {83.99} \\
     {25 image}& {74.68} & {75.85} & {84.33} & {85.60} \\
     {50 image}& {75.15} & {77.64} & {86.14} & {86.69} \\
     {100 image}& {76.33} & {80.14} & {87.71} & {90.01} \\
    
    \toprule
  \end{tabular}
\end{table}

\begin{table}[htpb]
      \centering
      \caption{Few-shot transfer with and without spatial augmentation for FF++ to AIF. }
      \label{tab:results_few_shot_aug_ff_to_aif}
  \begin{tabular}{c| c c| c c}
  \toprule
 {\textbf{Few-Shot}} & \multicolumn{2}{c|}{\textbf{Classifier}} & \multicolumn{2}{c}{\textbf{DDT (Ours)}} \\ 
 {\textbf{Images}} & {Without Aug} & {With Aug} & {Without Aug} & {With Aug} \\
\toprule
     {0 image}& {54.26} & {54.26} & {60.79} & {60.79}\\
     {5 images}& {55.31} & {56.75} & {64.21} & {67.82}\\
     {10 images}& {55.60} & {58.02} & {64.56} & {69.61}\\
     {25 image}& {56.79} & {60.65} & {66.76} & {70.11}\\
     {50 image}& {58.69} & {62.37} & {67.15} & { 69.56}\\
     {100 image}& {63.26} & {65.14} & {68.35} & {70.32}\\
    \toprule
  \end{tabular}
\end{table}

\begin{table}[htpb]
      \centering
      \caption{Few-shot transfer with and without spatial augmentation for FF++ to Dessa. }
      \label{tab:results_few_shot_aug_ff_to_dessa}
  \begin{tabular}{c| c c| c c}
  \toprule
 {\textbf{Few-Shot}} & \multicolumn{2}{c|}{\textbf{Classifier}} & \multicolumn{2}{c}{\textbf{DDT (Ours)}} \\ 
 {\textbf{Images}} & {Without Aug} & {With Aug} & {Without Aug} & {With Aug} \\
\toprule
     {0 image}& {64.88} & {64.88} & {74.28} & {74.28}\\
     {5 images}& {72.80} & {73.67} & {77.78} & {79.60}\\
     {10 images}& {73.35} & {74.20} & {78.94} & {81.18}\\
     {25 image}& {74.25} & {75.04} & {79.55} & {81.27}\\
     {50 image}& {76.00} & {77.05} & {79.54} & {82.91}\\
     {100 image}& {76.55} & {78.03} & {80.81} & {84.80}\\
    
    \toprule
  \end{tabular}
   
\end{table}

\end{appendix}

\end{document}